# On the Emergence of Syntax by Means of Local Interaction

**Zichao Wei**
Saarland University
ziwe00001@stud.uni-saarland.de

## Abstract

Can syntactic processing emerge spontaneously from purely local interaction? We present a concrete instance on a minimal system: an 18,658-parameter two-dimensional neural cellular automaton (NCA), supervised by nothing more than a 1-bit boundary signal, is trained on the membership problem of an arithmetic-expression grammar. After training, its internal $L \times L$ grid spontaneously self-organizes into an ordered, spatially extended representation that we name **Proto-CKY**. This representation satisfies three operational criteria for syntactic processing: expressive power beyond the regular languages, structural generalization beyond the training distribution, and an internal organization quantitatively aligned with grammatical structure (Pearson $r \approx 0.71$). It emerges independently on four context-free grammars and regenerates spontaneously after perturbation. Proto-CKY is functionally aligned with the CKY algorithm but formally distinct from it: it is a **physical prototype**, a concrete instantiation of a mathematical ideal on a physical substrate, and the systematic distance between the two carries information about the substrate itself.

## 1 Introduction

Neural language models produce syntactically well-formed language. This is an empirical fact [1]. But when we open these systems up, we do not find explicit structures isomorphic to the syntactic objects familiar from classical linguistics: non-terminals, derivation rules, and discrete parse trees do not appear in that form inside the weights and states [2], [3], [4]. Yet the outputs remain, to a substantial degree, syntactically legal.

LLMs let us see that this is possible, but they do not tell us how it happens. Their sheer scale is itself an obstacle: they are too large for us to locate a precise site at which "structured processing appears." Existing internal probes often carry enough computational capacity of their own that the "structure" they discover may reflect the probe's power rather than the model's representation [5], [6]. They cannot directly answer the question we want to ask. Let us therefore step back and pose the same question to a system simple enough to be observed in full: can a small, directly inspectable system also, on its own, handle a language that requires hierarchical structure? We do not treat it as a model of LLMs; we treat it only as an independent instance of the same phenomenon.

The strategy of this paper is to bypass natural language and start from a formal language with a precise mathematical description: an arithmetic-expression grammar. This grammar comes with a fully determined reference point: the CKY algorithm [7] gives a zero-error parse for it. Against this reference point we construct a minimal system: an 18,658-parameter two-dimensional neural cellular automaton [8], supervised only by a 1-bit boundary signal that spans the entire sequence.

What we observe is the spontaneous emergence of internal structure. The $L \times L$ grid self-organizes into an ordered, spatially extended representation. It satisfies the three operational criteria for syntactic processing that we state in the next section. It is quantitatively correlated with the CKY chart (Pearson $r \approx 0.71$), but it is not a copy of CKY: its states are continuous, it regenerates spontaneously after perturbation. It emerges independently on four distinct context-free grammars. We name this structure **Proto-CKY**.

This empirical observation is itself worth opening up. The body of the paper first defines the criteria (§2), then reports the observations and rules out alternative explanations one by one (§3–5). But all of this is groundwork. The real question comes at the end: **what, exactly, is the relation between the thing we have observed and the mathematical object it seems to point to?**

## 2 Background

The title of this paper promises "the emergence of syntax." Any careful reader will note that both words carry different meanings in different disciplinary traditions, and that in the overlapping public discourse these meanings are routinely conflated without scrutiny. Unless we give each concept an operational criterion that does not presuppose its conclusion, we are either forcing an ill-fitting label onto something or rigging the alignment.

### 2.1 Syntactic processing

If syntactic processing can emerge from a physical system, the emergent structure will almost certainly not be the kind of symbolic object that formal language theory takes as primitive. It will not be non-terminals, derivation rules, or discrete parse trees. To judge whether such a structure counts as "syntactic processing," we need a definition that does not presuppose symbolic form; otherwise we have already excluded every non-symbolic answer.

We adopt a behavioral working definition. It does not adjudicate the traditional disputes between formalist and functionalist approaches, between constituency and dependency; its commitments are only to observable behavior and inspectable internal organization. Concretely, we say that a system **exhibits syntactic processing** if it satisfies all three of the following:

1. **Expressive power beyond the regular languages.** The system correctly handles at least one class of phenomena expressible only by a CFG [9] (such as bracket matching or operator precedence), and its behavioral boundary clearly exceeds what any finite-state machine can reach.

2. **Structural productivity.** On inputs outside the training distribution, whether legal or illegal, the system exhibits systematic, structured response patterns rather than arbitrary or surface-statistics-driven ones. The spirit of this criterion traces back to Humboldt's classical observation about language as "the infinite use of finite means": a finite set of rules capable of producing infinitely many legal structures. In experiments, we test it through generalization. This criterion rules out lookup tables and shallow template matching.

3. **Syntactically relevant internal organization.** The internal states of the system spontaneously form an organization quantitatively aligned with grammatical structure. This criterion rules out black boxes that happen to produce correct outputs but have no internal structure, as well as systems that have internal organization unrelated to grammar.

**A note on these three.** We deliberately do not require perfect classification on all legal and illegal inputs. Consider an obvious example: humans. No one would deny that humans command syntax, yet humans are fooled by garden-path sentences [10], [11], break down on center-embedding at depth $\geq 3$ [12], [13], and disagree with one another on marginal grammaticality judgments [14]. That a physical system handling hierarchical structure exhibits boundary quirks of this kind does not mean it lacks syntactic processing. The three criteria demand structured productivity aligned with grammar, not zero error. In the experimental sections we report how well the system satisfies each criterion, including both its modes of success and its boundary biases.

The first two criteria are behavioral: they specify what input-output relations count as "getting it right." The third is representational: it requires that getting it right passes through some internal organization. Together the three constitute a falsifiable working definition. It is worth noting that the third criterion is difficult to verify directly for biological language processors (it requires neuroscientific tools), but for a controlled artificial system it is among the most directly observable.

### 2.2 Emergence

"Emergence" has a popular usage in the contemporary machine-learning literature: a model, upon scaling up, suddenly displays capabilities not explicitly taught during training. Under this usage, emergence is an empirical phenomenon about scale, making no commitment about the internal mechanisms that produce it.

This notion of emergence is not itself beyond question. Schaeffer et al. [15] showed that the supposed capability jumps may be measurement artifacts created by nonlinear evaluation metrics: switch the metric, and the jump vanishes. An "emergence" that can appear or disappear depending on the choice of yardstick is not a reliable scientific concept.

This paper takes emergence in its original sense [16]: many simple components interacting under local rules, without external coordination, spontaneously produce ordered structure at the global level. Emergence is not "unexpectedly getting it right"; it is "global order growing spontaneously from local interaction."

To sharpen this intuition into testable constraints, we say that a global property of a system satisfies the **emergence constraints** if the process that produces it meets all four of the following:

1. **Local connectivity.** The state at any position depends only on a fixed-size neighborhood, independent of the overall system size.
2. **Rule sharing.** All positions use the same update rule, independent of coordinates.
3. **No external coordination.** There is no external algorithm that decides the update order, and no global synchronization across neighborhoods.
4. **Simultaneous evolution.** All positions update in parallel at the same discrete time step; time is homogeneous.

Together the four characterize a centerless process: no global scheduler, no privileged position, no direct communication across neighborhoods. Any global order, if it appears, can only grow spontaneously from the iteration of local rules.

As a contrast, consider the existing approaches to syntactic processing. Whether classical parsing algorithms (CKY, the Earley parser, shift-reduce) or neural mechanisms (attention [17]), they all rely on some form of global coordination. Take CKY as an example: it fills a two-dimensional chart bottom-up, with each cell $(i, j)$ recording the set of non-terminals that can derive the substring $[i, j]$, yielding a zero-error answer. But the filling order is globally choreographed by span length; the chart itself has no dynamics. CKY is a **designed** solution, not an **emergent** one. In this paper it serves as a reference point: a known mathematical object with sharp symbolic meaning, against which we compare the emergent internal structure.

A system satisfying all four constraints is, mathematically, a cellular automaton (CA). This paper uses its neural variant, the neural cellular automaton (NCA), whose architecture is described in the next section. It is worth pointing out that the emergence constraints are stronger than "convolutional neural network": a deep convolutional network has a fixed number of layers that implicitly prescribes the rounds of information propagation. The emergence constraints require a single local rule applied repeatedly until convergence, with no architectural partitioning into rounds.

The experimental question of this paper can now be stated precisely: **on a system satisfying the emergence constraints, can ordered structure arise spontaneously to perform syntactic processing that satisfies all three criteria?**

### 2.3 Relation to prior work

This question connects to three existing lines of research.

The first is **linguistic structure inside neural networks.** That neural networks implicitly acquire representations correlated with linguistic structure is an observation with a clear lineage. Manning et al. [18] surveyed the evidence that self-supervised training yields language-like structure inside ANNs; Lake and Baroni [19] tightened the pressure test on compositional generalization; McCoy et al. [20] warned that "getting it right" does not always mean "getting it right for the right reasons." We share the source of this question (neural systems perform structured processing, but in what form?) while shrinking it to a scale at which the inside can be directly inspected, rather than measuring behavioral signatures on a large model.

The second is **the expressive power of neural networks on formal languages**: which classes of language RNNs and Transformers can recognize, how much memory they require, and where the boundary of Turing completeness lies [21], [22], [23], [24], [25]. For the cellular automaton used in this paper, the expressive-power question is trivial: CA has been proven Turing-complete [26] and can in principle compute any computable function. But Turing completeness only answers "can it?", not "in what form?" This paper is concerned with the latter: when a Turing-complete local system succeeds on a syntactic task, what does the internal structure it gives rise to look like? We inherit the methodology of this tradition (using formal languages as a controlled test bed), but the question itself lies off the expressive-power track.

The third is **neural cellular automata.** Since Mordvintsev et al. [8] first introduced them, NCAs have been applied to 3D construction [27], self-classification [28], control tasks [29], and attention-augmented variants [30]. The present paper is the first application of NCAs to the membership problem on a formal language.

## 3 Experimental Setup

The experimental task must satisfy two conditions: it must genuinely require hierarchical analysis (otherwise the first criterion for syntactic processing cannot be tested), and it must be simple enough to control and interpret. We choose an arithmetic-expression grammar. It shares the core syntactic features of natural language (hierarchical constituent structure, recursive nesting) without the additional complexity introduced by semantic and pragmatic factors. Concretely, the membership problem is defined on the following grammar:

$$E \to E + T \mid T, \quad T \to T \times F \mid F, \quad F \to (E) \mid \mathrm{id} \tag{1}$$

Converted to Chomsky normal form this yields 5 tokens, 11 non-terminals, and 9 binary rules. Given a token sequence, the model decides whether it is a legal expression. The grammar has operator precedence and recursive nesting; its CKY chart is non-trivial.

The model is a 2D NCA defined on an $L \times L$ grid, with $C = 2$ state channels and 18,658 parameters. The grid is indexed by the matrix convention: cell $(i, j)$ lies in the $i$-th row (corresponding to the $i$-th token of the sequence) and the $j$-th column (corresponding to the $j$-th token). At $t = 0$ the grid is completely uniform (all zeros). The update rule is:

$$\boldsymbol{h}_{t+1}(i,j) = \sigma\big(W_h \cdot \mathrm{relu}\big(W_2 \cdot \mathrm{relu}\big(W_1 \cdot \big[\boldsymbol{e}_i; \boldsymbol{e}_j; W_s \boldsymbol{h}_t(i,j)\big]\big)\big)\big) \tag{2}$$

where $\boldsymbol{e}_i, \boldsymbol{e}_j \in \mathbb{R}^d$ are the learned embeddings of the $i$-th and $j$-th tokens ($d = 16$), $W_s$ projects the current state into the embedding space, $W_1, W_2$ are $3 \times 3$ convolution kernels, $W_h$ is a $1 \times 1$ output head, and $\sigma$ is the sigmoid. Token information is injected externally at every step; it is not stored on the grid.

This architecture satisfies the four emergence constraints stated above: the receptive field is a $3 \times 3$ neighborhood (local connectivity); $W_1, W_2, W_h$ are shared across all spatial positions and all time steps (rule sharing); no external algorithm decides the update order (no external coordination); all cells update in parallel at the same discrete step (simultaneous evolution). From the perspective of rule-execution dynamics, this is a standard cellular automaton; the only difference is that the rule is learned rather than hand-designed.

The supervision signal is only 1 bit: the upper-right cell $(0, L-1)$, the position that spans the entire sequence, has its channel 0 trained by BCE loss to predict sequence legality. The remaining $L^2 \times C - 1$ grid values receive no supervision whatsoever. Training data is generated on the fly (batch size 64), with sequence length $L \leq 12$ and balanced positive and negative samples. Negative samples are constructed by four strategies: random token sequences, mismatched parentheses, double operators, and missing operators. For each training sample, the number of NCA iterations is sampled uniformly from $[3, \max(6, 2L)]$, forcing the model to produce correct predictions at varying iteration depths. The optimizer is Adam (learning rate $10^{-3}$); training converges within roughly 2000 gradient steps.

At test time the NCA iterates until convergence ($\max|\boldsymbol{h}_{t+1} - \boldsymbol{h}_t| < \varepsilon = 0.01$), with a step cap of $\max(50, L)$, generous for short sequences and scaling linearly with length for long ones. OOD evaluation uses a balanced protocol: at each test length, 100 legal flat expressions and 100 constructively illegal sequences, avoiding dependence on a CKY verifier so that evaluation is possible at arbitrary length.

To probe universality, we train the same architecture on five additional languages (varying only the training data): Dyck-1 (matched parentheses, 2 tokens), Dyck-2 (two kinds of parentheses, 4 tokens) [31], NL Agreement (subject-verb agreement, 7 tokens) [32], [33], three context-free grammars, plus $a^*$ (all-$a$ sequences) and $a^*b^*$ ($a$'s followed by $b$'s), two regular languages serving as negative controls.

## 4 Experiments

### 4.1 The emergence of syntax

We have laid down three criteria for syntactic processing: expressive power beyond the regular languages, systematic productivity, and syntactically relevant internal organization. We have also established that our model is a system driven by local interaction alone. The question now is: can such a system satisfy all three at once?

Training converges within about 2000 steps. The converged model reaches 100% classification accuracy on the training distribution ($L \leq 12$). The membership problem for arithmetic expressions genuinely requires the expressive power of a context-free grammar; in this sense, the first criterion is met.

The second criterion demands systematic productivity. If the model had merely memorized the statistical regularities of the training distribution, it should fail rapidly outside that distribution. In fact, on sequences far beyond the training length ($L = 50, 100, 500$, all the way to $L = 1000$, more than 80 times the training length), the model maintains 100% accuracy (Figure 1).

A detail worth explaining: training samples are generated on the fly within the window $L \leq 12$, and in that window the overwhelming majority of randomly sampled legal expressions are flat. Expressions of nesting depth $\geq 2$ account for only a small fraction of the training set. The main model therefore generalizes in the flat direction to very long sequences, but on purely nested inputs of depth $\geq 2$ it fails from the start. This is not an architectural defect but a bias of the training distribution. When we explicitly include depth 2–4 nested expressions at roughly 30% of training samples (deep augmentation), the same architecture generalizes to depth 10 on pure nested inputs and depth 7 on mixed depth-length inputs (Figure 2).

There is a trade-off between length generalization and depth generalization: under a single set of weights, strengthening depth generalization weakens length generalization, and vice versa. This trade-off also appears on a Transformer baseline of comparable scale (Appendix A); it is not peculiar to the NCA.

Regardless of which direction one looks, a structured mechanism operating stably outside the training distribution has emerged: past the boundary of any finite-state machine in the length direction, and into nesting structures never seen during training in the depth direction. The second criterion is met.

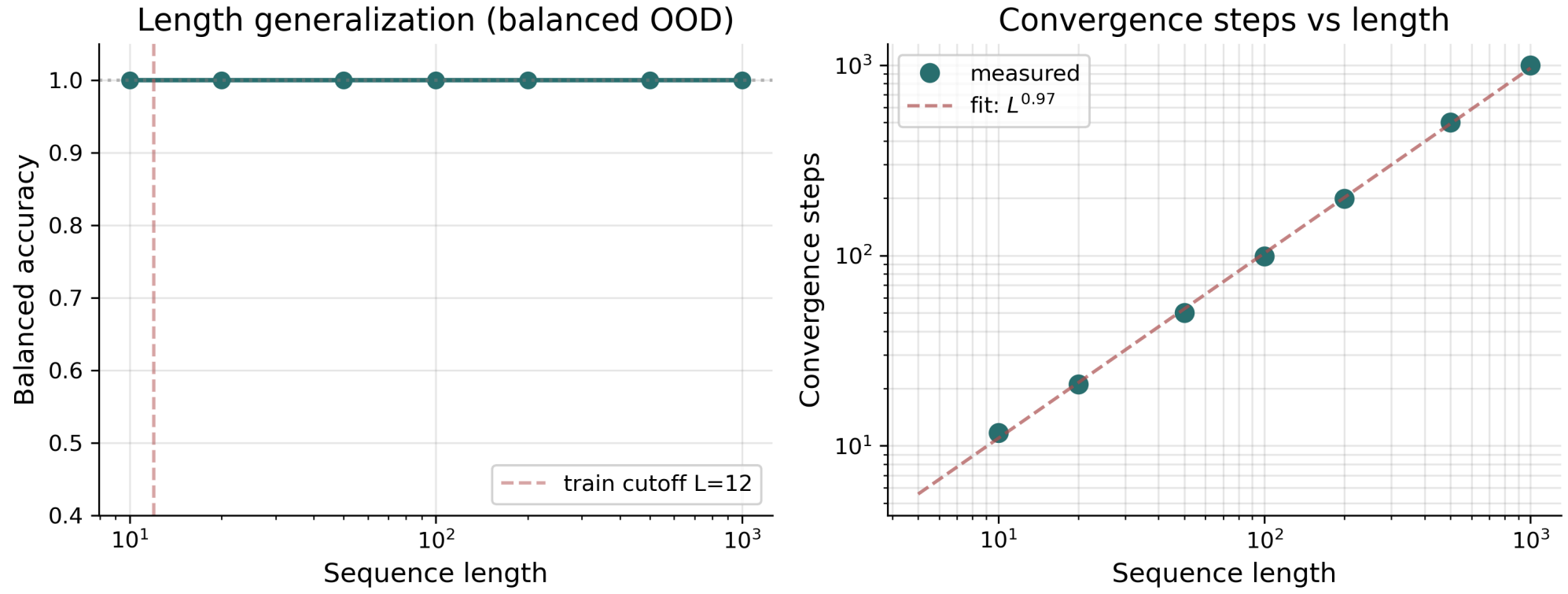

Figure 1: Length generalization. Left: balanced OOD accuracy across seven lengths from the training regime ($L \leq 12$) through $L = 1000$; accuracy stays at 100% across the entire range. Right: convergence steps scale near-linearly with sequence length.

Depth generalization

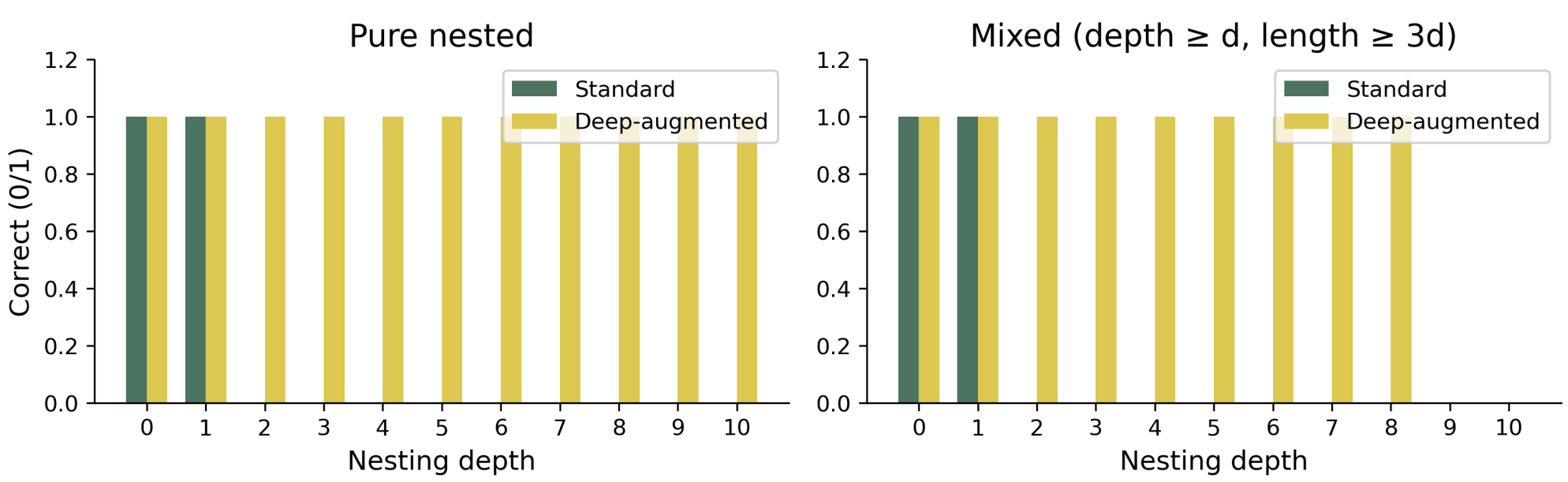

Figure 2: Depth generalization on pure-nested (left) and mixed depth-length (right) inputs, for the main model (Standard) and the deep-augmented variant (Deep-augmented). The main model handles only depths 0 and 1; the deep-augmented variant reaches depth 10 on pure nested and depth 7 on mixed. Deep augmentation recovers depth generalization without changing the architecture.

The third criterion is the most important, and the hardest. It requires that getting it right passes through some internal organization. An input-output black box, even if it satisfies the first two criteria, does not count as syntactic processing: the essence of syntax is processing of structure, not merely reacting to structured inputs. We must open the model up.

Before inference begins, the $L \times L$ grid is completely uniform: all cells are zero, carrying no information about the input, let alone any structure. After iterative relaxation, the grid self-organizes into a highly ordered activation pattern. This fact deserves emphasis: $L^2 \times C - 1$ grid values receive no supervision at all; their ordered organization is entirely spontaneous.

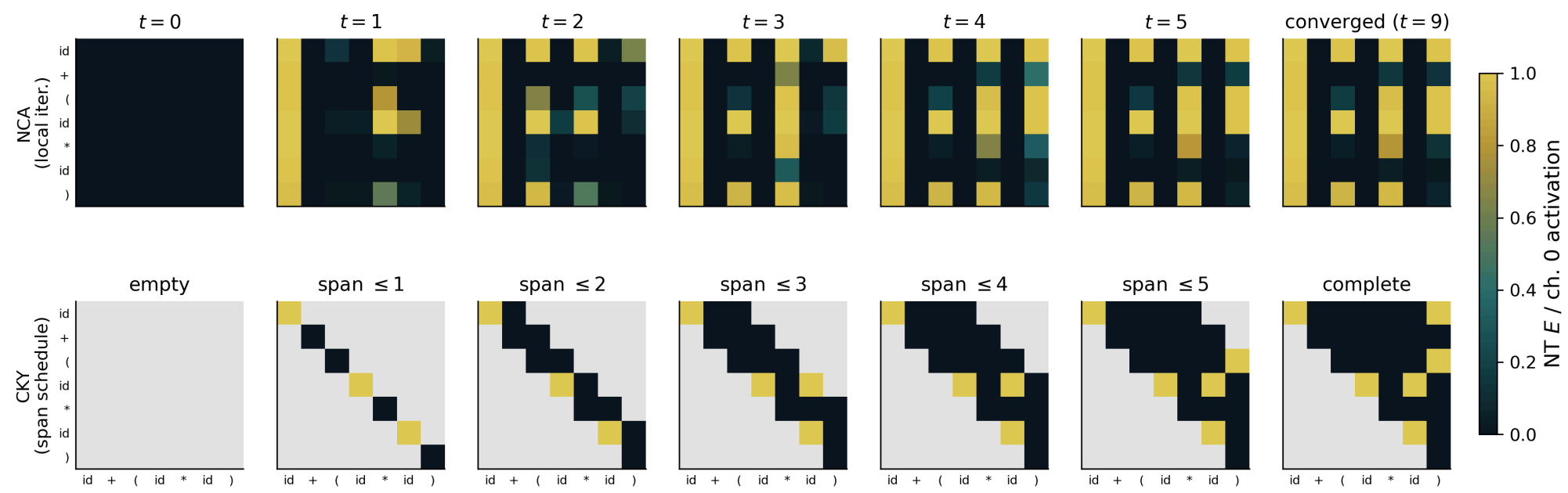

Figure 3: Top row: NCA grid (channel 0) at iteration steps $t = 1..5$ and after convergence, from a zero initialization. Bottom row: CKY chart (NT $E$ indicator) revealed incrementally by span length — masked cells mark spans not yet computed. Both processes fill the upper-triangular chart from smallest spans outward; NCA does so by local wavefront, CKY by an external span-length schedule. Columns align at corresponding progression positions, not at identical step counts.

We name this ordered representation **Proto-CKY** (Figure 3). It shares a geometry with the CKY chart: both organize token pairs on a span lattice. But its internal mechanism differs substantially from CKY's. CKY fills the chart by discrete enumeration; Proto-CKY converges on the grid by continuous relaxation. CKY is choreographed by an external algorithm proceeding in order of span length; Proto-CKY is produced by a single local rule applied repeatedly until it reaches an attractor. The prefix "Proto-" does not refer to a "pre-syntactic stage" in the evolutionary-linguistics sense. The judgment behind the name is deferred to the Discussion. For now, treat it as a descriptive label.

At this point, all three criteria are satisfied. But satisfying them alone does not complete the argument. We must also rule out several ways in which Proto-CKY might fail to count as "syntactically relevant internal organization." That is the task of the next three sections.

### 4.2 Proto-CKY is not noise

A natural suspicion is that the ordered representation is merely a byproduct that any 2D NCA architecture would produce on any sequence-classification task. If so, it should appear indiscriminately across tasks, regardless of whether they involve hierarchical structure.

The most direct test is to find tasks that do not require syntax. We train the same architecture on two regular languages: $a^*$ (all-$a$ sequences) and $a^*b^*$ ($a$'s followed by $b$'s). Both models successfully learn their respective tasks, but both grids remain flat: the variance of the internal representation is $1.5 \times 10^{-3}$ and $4.6 \times 10^{-3}$, respectively, roughly two orders of magnitude below the 0.194 measured on the arithmetic grammar. The decision rules for these tasks are minimal ("does each position satisfy a local condition?"), and the models correspondingly develop no appreciable internal organization. The ordered representation is not a default behavior of NCAs on arbitrary tasks.

Conversely, if the ordered representation is related to syntax, it should recur across different syntactic tasks. We train the same architecture separately on three other context-free grammars, Dyck-1 (matched parentheses), Dyck-2 (two kinds of parentheses), and NL Agreement (subject-verb agreement), varying only the training data and no part of the architecture. All four grammars independently produce ordered representations (Figure 4). Four entirely different grammars, one local rule, and a similar internal organization emerges spontaneously on every one of them.

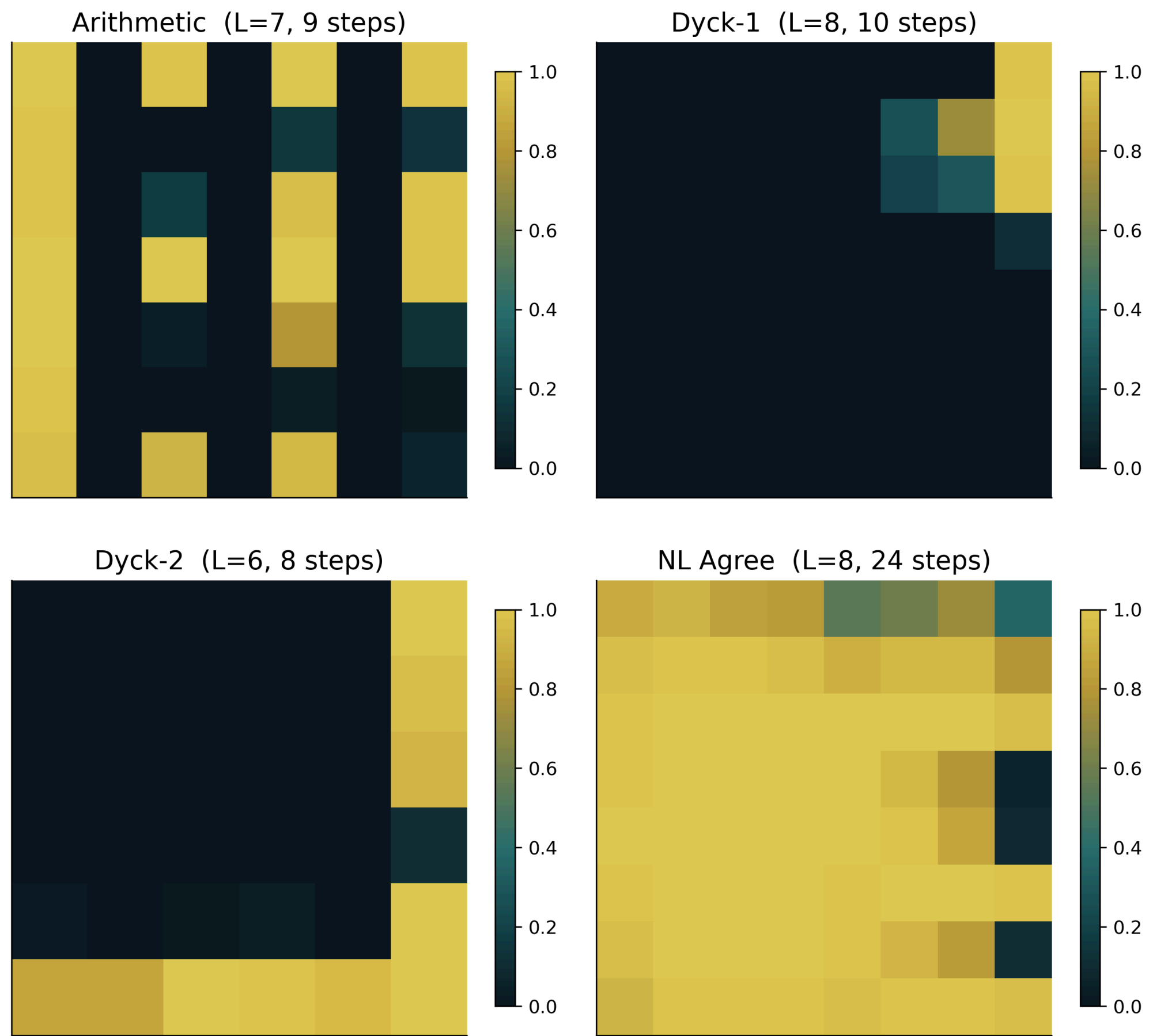

Figure 4: Converged NCA grids for the four context-free grammars (arithmetic, Dyck-1, Dyck-2, NL Agreement), all sharing the same architecture and training procedure and differing only in training data. The ordered representation emerges independently on each grammar.

Table 1 presents the results across all six languages side by side. Grid variance exhibits a two-order-of-magnitude gap between the regular languages and the CFGs. The internal activations on the four CFGs further show systematic positive correlation with their respective CKY charts ($r \approx 0.2 - 0.7$). The ordered representation is not a default behavior of the NCA; it appears only on tasks that require hierarchical analysis, and it is quantitatively aligned with grammatical structure.

Table 1: Internal organization across six languages. Grid variance measures whether the NCA develops non-trivial internal structure (upper-triangular activation variance, averaged over 100 legal samples). Pearson $r$ measures alignment between that structure and the CKY chart (correlation between channel 0 and the start-symbol indicator, aggregated over 200 samples). The two regular languages have variance below 0.005; the four CFGs have variance above 0.1; there are no intermediate values.

| **Language** | **Type** | **Grid variance** | **Pearson $r$** |
|---|---|---|---|
| $a^*$ | Regular | 0.003 | — |
| $a^*b^*$ | Regular | 0.005 | — |
| Dyck-1 | CFG | 0.101 | 0.22 |
| Dyck-2 | CFG | 0.145 | 0.25 |
| Arithmetic | CFG | 0.191 | 0.71 |
| NL Agreement | CFG | 0.204 | 0.22 |

### 4.3 Proto-CKY is not a copy of CKY

Since Proto-CKY is functionally aligned with the CKY chart, one might ask: has the NCA simply learned to imitate the CKY algorithm internally? If so, the emergence would be less striking; it would be nothing more than a learned clone of a known algorithm.

Let us check directly. If the NCA were copying CKY, its internal activations should be highly consistent with the CKY chart, and we should expect a correlation coefficient near 1. The measured Pearson correlation is $r = 0.709$ (for the start non-terminal $E$, on the representative expression `id + id * id`; the randomly-initialized baseline is $r \approx -0.087$, standard deviation 0.077). There is systematic positive correlation, but $r = 0.71$ is far from a copy. The two correspond, but they are plainly not the same thing (Figure 5).

A more intuitive observation comes from inspecting the converged grid. In the CKY chart, each cell $(i, j)$ precisely records the set of non-terminals that can derive the substring $[i, j]$; different cells can be entirely different. Proto-CKY has a markedly different morphology: the converged grid looks more like a textured continuous field than a discrete lookup table. The organizational principles of this structure are not yet fully understood.

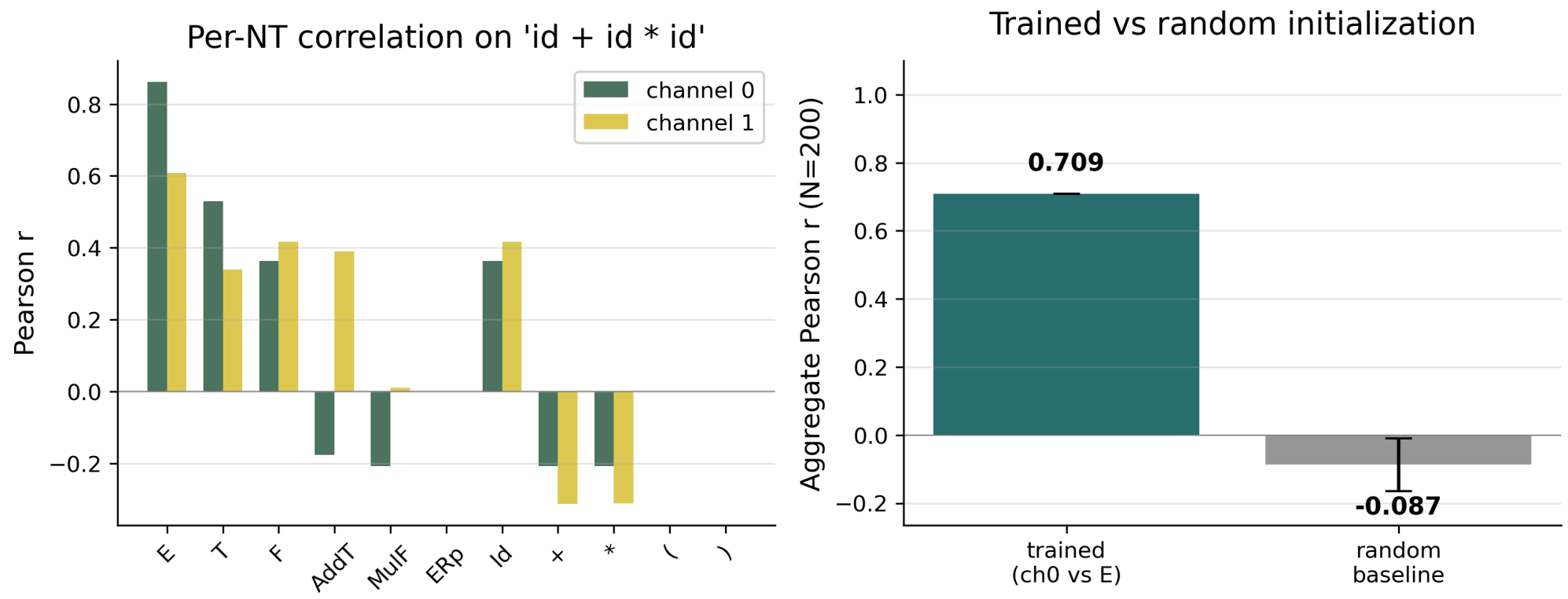

Figure 5: Left: Pearson $r$ between NCA channel activations and CKY-chart non-terminal indicators on the representative expression `id + id * id`, broken down by non-terminal. Right: aggregate $r$ between channel 0 and the E non-terminal on 200 samples, compared to a randomly-initialized NCA baseline. The aggregate correlation is well above random but far below unity.

Our understanding of Proto-CKY's internal structure remains rather limited. But one thing is clear: it is not a copy of CKY. It shares CKY's function, organizing tokens into hierarchical constituents, but it does so through a different and not yet fully understood internal implementation. That two fundamentally different computational mechanisms, discrete enumeration and continuous relaxation, converge functionally is itself a striking phenomenon.

### 4.4 Proto-CKY is not a coincidence

So far we have ruled out two alternative explanations: Proto-CKY is not architectural noise, and it is not a copy of CKY. But it could still be fragile, a coincidence that happens to appear under a particular random seed or particular hyperparameter setting.

Consider the simplest robustness check. We train the same architecture on 10 different random seeds. All 10 runs produce ordered representations; 9 of them reach 100% balanced OOD accuracy at $L = 500$, and the tenth reaches 93.3%. The Pearson $r$ measured per seed has a mean of 0.60 (standard deviation 0.10), consistent with the $r = 0.71$ observed for the single seed. In a capacity ablation, reducing the hidden dimension from $d = 16$ to $d = 4$ (shrinking the parameter count from 18,658 to 1,210, a 15-fold reduction) preserves 100% OOD accuracy at $L = 500$, and the ordered representation still appears. Proto-CKY is not picky about training conditions (Table 2).

Table 2: Robustness across random seeds and model capacities. Top: 10-seed stability at the main $d = 16$ configuration; the worst seed reaches 93.3% on $L = 500$ OOD, the other nine reach 100%. Bottom: single-seed runs across hidden dimensions; the model remains stable down to roughly 1/15 of the main parameter count.

| Setting | Params | In-dist | $L = 500$ OOD | Pearson $r$ |
|---|---|---|---|---|
| *Seed stability* ($d = 16$, 10 seeds) | | | | |
| Best seed | 18,658 | 100% | 100% | – |
| Worst seed | 18,658 | 100% | 93.3% | – |
| Mean ± std | 18,658 | 99.98 ± 0.06% | 99.33 ± 2.11% | 0.60 ± 0.10 |
| *Capacity ablation* (single seed) | | | | |
| $d = 4$ | 1,210 | 97.4% | 100% | – |
| $d = 8$ | 4,722 | 99.6% | 100% | – |
| $d = 16$ (main) | 18,658 | 100% | 100% | 0.71 |
| $d = 32$ | 74,178 | 100% | 100% | – |

Injecting Gaussian noise ($\sigma = 1.0$) into the grid at an arbitrary step during inference, or resetting the entire state to zero, the model recovers to 100% accuracy in every case, at the cost of additional convergence steps. Structure, once destroyed, spontaneously regenerates. This is not the behavior of a fragile artifact: fragile things, once broken, do not regrow on their own (Figure 6).

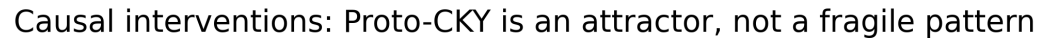

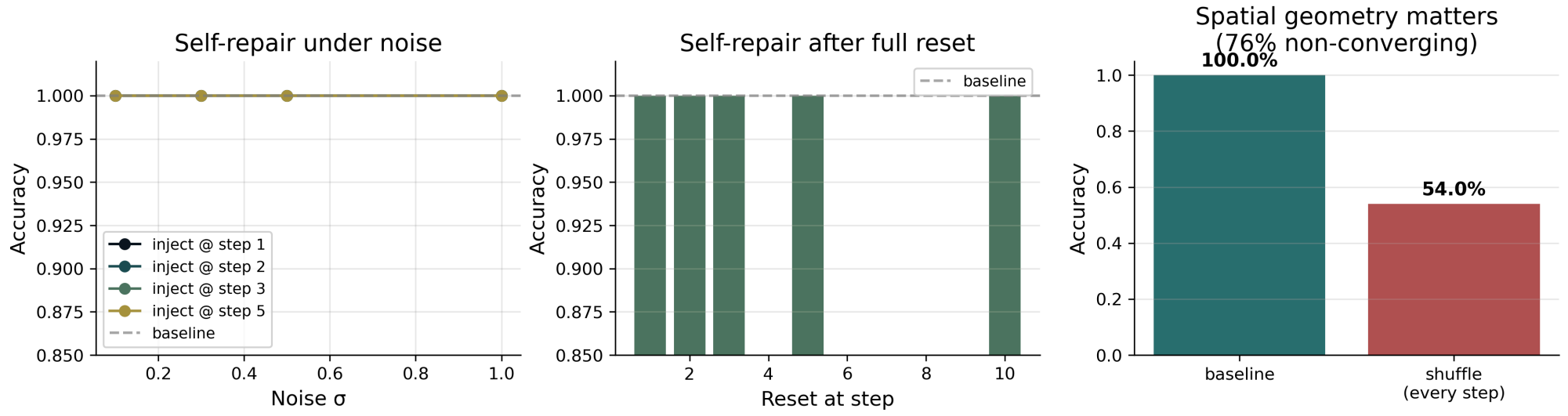

Figure 6: Causal interventions. Injecting Gaussian noise at various inference steps, resetting the state to zero, and region freezing all recover full accuracy (with extra convergence steps), while spatial shuffling every step collapses the dynamics. Structure is regenerated from local rules once the geometry is intact.

Further intervention experiments reveal what Proto-CKY depends on. Freezing 70% of the grid cells (randomly chosen) drops accuracy from 100% to 85%; freezing 50% still yields 92%; freezing 30% still yields 99%. Most cells are not irreplaceable parts; they are redundant. But randomly shuffling the spatial positions of all cells at every inference step (keeping all values intact, only changing their relative positions on the grid) collapses accuracy from 100% to 54%, with 76% of samples failing to converge within the iteration cap. Proto-CKY's function depends on its spatial organization, not on the values of individual cells within that organization. Values can be replaced or perturbed; their relative positions on the grid cannot be scrambled. Table 3 consolidates the full intervention sweep.

Table 3: Causal interventions during inference. Noise injection, state reset, and single spatial shuffles all recover; sustained shuffling collapses the dynamics. Region freezing degrades gracefully with the fraction frozen; the upper-triangle freeze (which disables chart-area computation) yields accuracy comparable to random 70% freezing, suggesting an empirical heuristic floor of about 85% on this task.

| **Intervention** | **Configuration** | **Accuracy** | **Note** |
|---|---|---|---|
| Noise injection | $\sigma \in [0.1, 1.0]$, at step 1–5 | 100% | recovers |
| State reset | reset at step 1, 2, 3, 5, 10 | 100% | recovers |
| Spatial shuffle (single) | at step 1, 2, 3, or 5 | 100% | recovers |
| Spatial shuffle (sustained) | at every step | 54.0% | 76% non-conv. |
| Region freeze | random 30% of cells | 98.7% | |
| Region freeze | random 50% of cells | 92.3% | |
| Region freeze | random 70% of cells | 84.7% | |
| Region freeze | random 90% of cells | 60.3% | |
| Region freeze | upper triangle (chart area) | 85.3% | heuristic floor |
| Region freeze | lower triangle (non-chart) | 100% | |
| Region freeze | diagonal | 92.7% | |

## 5 Discussion

### 5.1 What have we observed?

Let us step back and ask this question.

In the preceding sections we observed an ordered, spontaneously emerging internal representation. It satisfies three operational criteria; it is quantitatively correlated with the CKY chart without being equal to it; it appears stably across seeds, grammars, and capacities; it self-repairs after perturbation. We gave it a name: Proto-CKY. **But from an ontological standpoint, what kind of thing have we actually observed?**

Proto-CKY possesses properties that no mathematical object can have: it is continuous, finite-precision, self-repairing, and still functional when most of its cells are frozen. These are properties of a physical process, not of a symbolic system. This forces us to confront a more primitive question: **what is "syntax"?** In the conventions of formal language theory, "syntax" names a symbolic object: context-free grammars, derivation rules, parse trees, the CKY algorithm. It is zero-error, unbounded in capacity, and independent of any physical substrate. It is mathematically complete; its "existence" does not depend on any particular implementation.

But the word "syntax" has another use. When we say "humans command syntax," we plainly do not mean a mathematical object. We mean a processing activity that takes place on a physical substrate: finite-capacity, fooled by garden-paths, collapsing at depth 3 or more, varying among individuals on marginal grammaticality. What humans "command" is not the object in Plato's heaven. It is a physical process, one broadly aligned in function with that mathematical object but retaining a distance peculiar to physical realization.

The word "syntax" has in fact always carried two ontologically distinct meanings: one a mathematical object, the other a physical process. The two are rarely distinguished explicitly, because in the formal-language tradition the mathematical object is taken by default as the primary subject of study, and physical realization is treated as "engineering detail." But when we confront an object that has emerged spontaneously from local rules, this default ordering no longer holds. What emerges is, by definition, not and cannot be a mathematical object; it is physical. It cannot satisfy any of the three conditions: zero error, unbounded capacity, substrate-independence.

Proto-CKY belongs to the latter category. It is not an instance of "syntax" in the mathematical-object sense. It belongs to the family where "humans command syntax" also sits: a physical process, carrying all the hallmarks of physical realization. We named it Proto-CKY because its relation to CKY is that of a physical prototype to a mathematical ideal: aligned in function, separated by a measurable distance in form.

This distance deserves to be taken seriously. If Proto-CKY were merely a numerical approximation to CKY (the same function realized at finite precision), all its properties should follow CKY's. Self-repair should not appear: a numerical approximation, once broken, does not regrow. Redundancy should not appear: every cell of CKY is independently computed, with no redundancy. Proto-CKY exhibits precisely these deviations. They tell us it is not a shadow of CKY but a different class of object: a physical process with its own structure, one that happens to align functionally with CKY. For a given mathematical ideal, what does its physical prototype look like? This question has an entire tradition in physics. In linguistics, this distinction has not been handled with comparable rigor: formalized grammars and the language processing that actually takes place on physical substrates are often taken by default to be the same object. Proto-CKY makes the distinction visible in a concrete case.

### 5.2 Geometry and dynamics

Why does Proto-CKY arise here? We cannot give a complete answer, but we can point to two necessary ingredients.

The first comes from the geometry of the 2D grid. Under a $3 \times 3$ receptive field, cell $(i, j)$ is the position that, along the shortest path, can simultaneously receive influence from token $i$ and token $j$. The upper triangle therefore becomes the natural coordinate system for token-pair interaction under this locality constraint. At the same time, the readout position $(0, L - 1)$ can only access its $3 \times 3$ neighborhood. For the 1-bit supervision signal to depend on global information, the model must organize some intermediate representation on the interior grid to relay distant information. Geometry provides a natural coordinate system; local supervision forces information to traverse the grid. Together they form an inductive bias toward span-like organization.

But geometry explains only the coordinate system, not the content. Why does the system converge to an ordered state along this coordinate system? This is the second ingredient: dynamics. The self-repair experiments suggest that Proto-CKY is not a static pattern that training has "placed" in the right position; it behaves more like an attractor of a dynamical system, pulled back when displaced. The collapse under spatial shuffling further shows that the existence of this attractor depends on correct spatial structure: values can be broken, and the system recovers; but once the geometry is scrambled, the attractor itself ceases to exist. Geometry provides the coordinate system, dynamics drives convergence; neither suffices without the other. As for the specific mechanism by which a local rule produces global order through iteration, we candidly acknowledge that we do not yet have a clear theoretical framework to explain this process.

### 5.3 Open questions

Even granting the geometry-plus-dynamics framework, many phenomena remain unexplained.

The internal organizational principles of Proto-CKY are the most conspicuous open question. The converged grid displays an ordered continuous-field-like structure. Visually, one can observe a tendency toward column-wise organization, but this tendency weakens as input length grows, and the underlying mechanism remains unclear. More fundamentally: what are Proto-CKY's cells encoding? Each cell of the CKY chart has a clear symbolic meaning (a set of non-terminals). Proto-CKY's cells do not.

The CKY chart uses only the upper triangle ($i \leq j$). The NCA grid has no such constraint; the lower triangle also participates in updates and information propagation. What role does the lower triangle play in Proto-CKY's function? More interestingly, the activation pattern of the lower triangle is symmetric with that of the upper triangle, yet local-freezing experiments suggest the lower triangle is in fact unnecessary for function. Is it a redundant byproduct of the propagation process? An alternative path occasionally exploited on certain samples? Or do the lower-triangle activations carry, from the model's perspective, a physical referent different from what a CKY chart expresses?

Finally, and perhaps most fundamentally: Proto-CKY's constellation of properties (self-repair under noise, robustness to local freezing, fragility under spatial shuffling) evokes, in the physical picture, something closer to a crystal than to a clock. A clock's order comes from deliberate engineering, with every gear playing a precise causal role. A crystal's order comes from physical conditions: introduce impurities, and it is largely unchanged; heat it, and it melts; cool it, and it recrystallizes; but alter the lattice constant, and the same crystal will not form. This is not a rigorous theory, but it suggests a direction: perhaps the right question is not "what does each cell of Proto-CKY do?" but "what conditions make this kind of order inevitable?"

## 6 Conclusion

**Structured, productive sequence processing can emerge spontaneously under minimal physical constraints [16]. The form it takes, at least in the one instance we have observed, is a physical prototype, not a mathematical object.**

Proto-CKY emerges under the constraints of 18,658 parameters, 1 bit of boundary supervision, and purely local interaction. It satisfies three operational criteria for syntactic processing. It appears independently on four context-free grammars and is absent on regular languages. It is quantitatively correlated with the CKY chart in function but maintains a distance in form peculiar to physical realization. This distance is not approximation error; it is information about the physical substrate itself.

Taken as a starting point, Proto-CKY turns an object that would otherwise be difficult to observe directly, namely **the form in which structured sequence processing emerges in local-rule systems**, into a concrete instance that can be opened in full, inspected cell by cell, and systematically ablated. On such an instance, many questions can now be asked meaningfully: what its internal mechanism specifically is; whether it generalizes to other families of formal languages; whether there are comparable properties between it and biological language processing; whether it is of the same kind as what happens inside large-scale neural language models. This paper does not answer these questions, but they can now be asked of a concrete, inspectable object.

Syntax in the physical sense, as an object of study, is something quite different from symbolic syntax. What this paper reports is one instance of it.

# A Appendix A: A contrast on a different architecture

The main paper reports Proto-CKY as a physical prototype emerging under minimal physical constraints: a continuous field organized along a span lattice that self-repairs after perturbation. A natural contrast is: what does the same task look like on an architecturally **entirely different** system, a Transformer? Does it develop the same geometry, or does it take an altogether different path?

This appendix reports parallel experiments on a Transformer baseline. The main conclusions fall into three layers: (i) with enough parameters, the Transformer achieves length generalization, but it follows a path sharply different from the NCA's, relying on the aggregation mechanism known as an attention sink rather than any spatial geometry; (ii) on depth generalization, Transformer and NCA behave alike: under standard training both handle only flat inputs, and deep augmentation lets both handle nested inputs but at a cost to length generalization; (iii) the length-vs-depth trade-off holds across both architectural families, so this limitation is more likely a property of the training protocol itself than of any particular architectural choice.

### A.1 A scale threshold

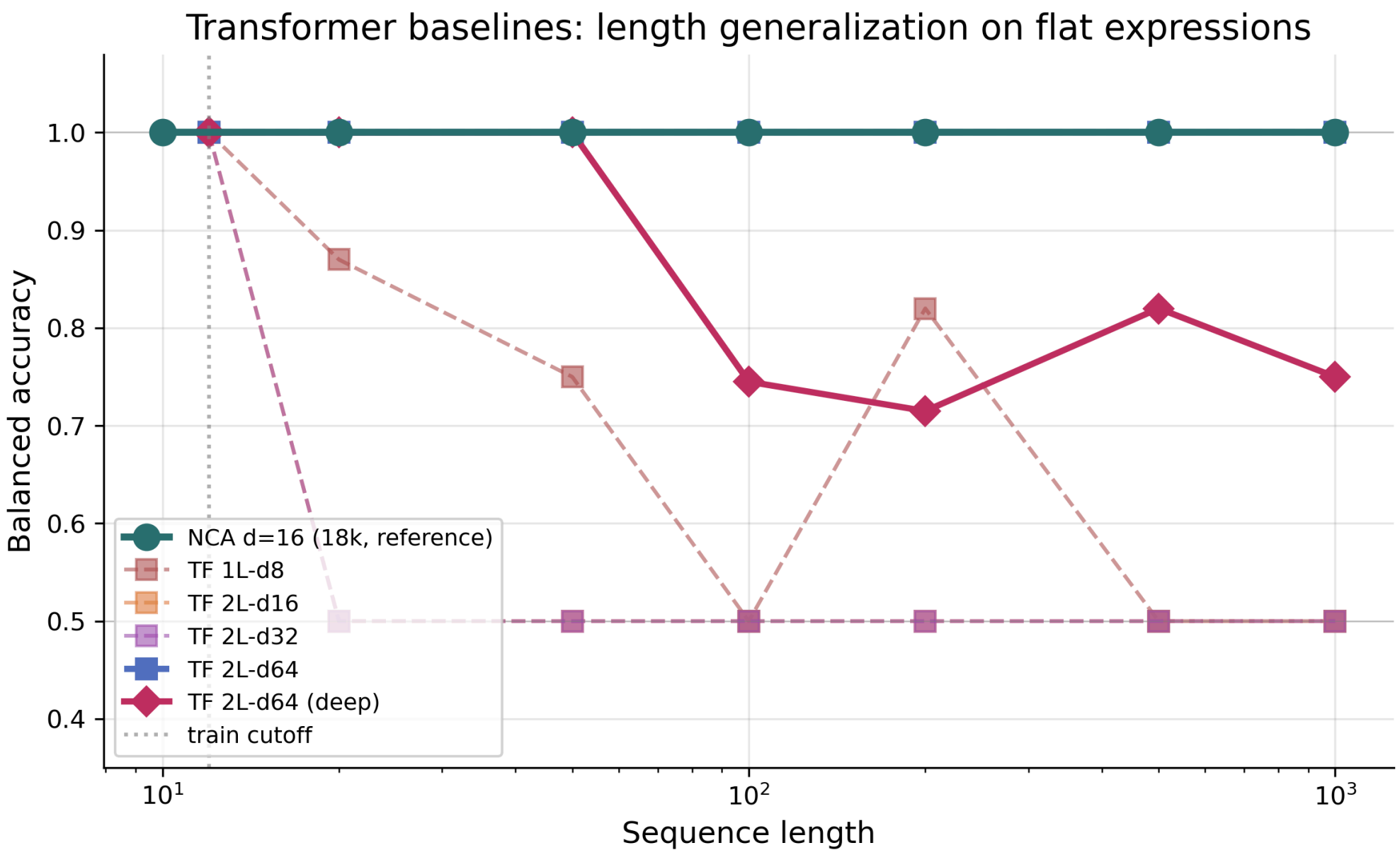

Figure 7: Length generalization of four Transformer configurations, compared to the main NCA (d=16, 18,658 parameters). All four Transformers reach 100% at $L \leq 12$ (train). In OOD, d=8, d=16, d=32 collapse to near chance by $L = 50$; d=64 (165,953 parameters, $\sim 9\times$ NCA) generalizes perfectly to $L = 1000$.

We train four Transformer configurations as baselines: 1-layer $d = 8$ (9,121 parameters), 2-layer $d = 16$ (23,057 parameters), 2-layer $d = 32$ (58,401 parameters), and 2-layer $d = 64$ (165,953 parameters). All four reach 100% on the training distribution $L \leq 12$, same as the NCA. Beyond the training length, however, their fates diverge sharply (Figure 7).

The first three Transformers ($d = 8$, $d = 16$, $d = 32$) collapse to near-chance by $L = 50$. Even $d = 32$, at 58k parameters (roughly 3 times the NCA), does not survive. Only at $d = 64$ (roughly 9 times the NCA) does length generalization abruptly appear, holding all the way to $L = 1000$. Between 58k and 166k parameters, the Transformer jumps from "no generalization at all" to "perfect generalization"; somewhere in that narrow interval a capability phase transition takes place.

The NCA crosses this phase transition already at 18k parameters, nearly an order of magnitude below the Transformer's threshold. But "parameter efficiency" is only the surface of the story. The question worth asking next is: is what $d = 64$ has achieved the same thing?

### A.2 Attention is not local

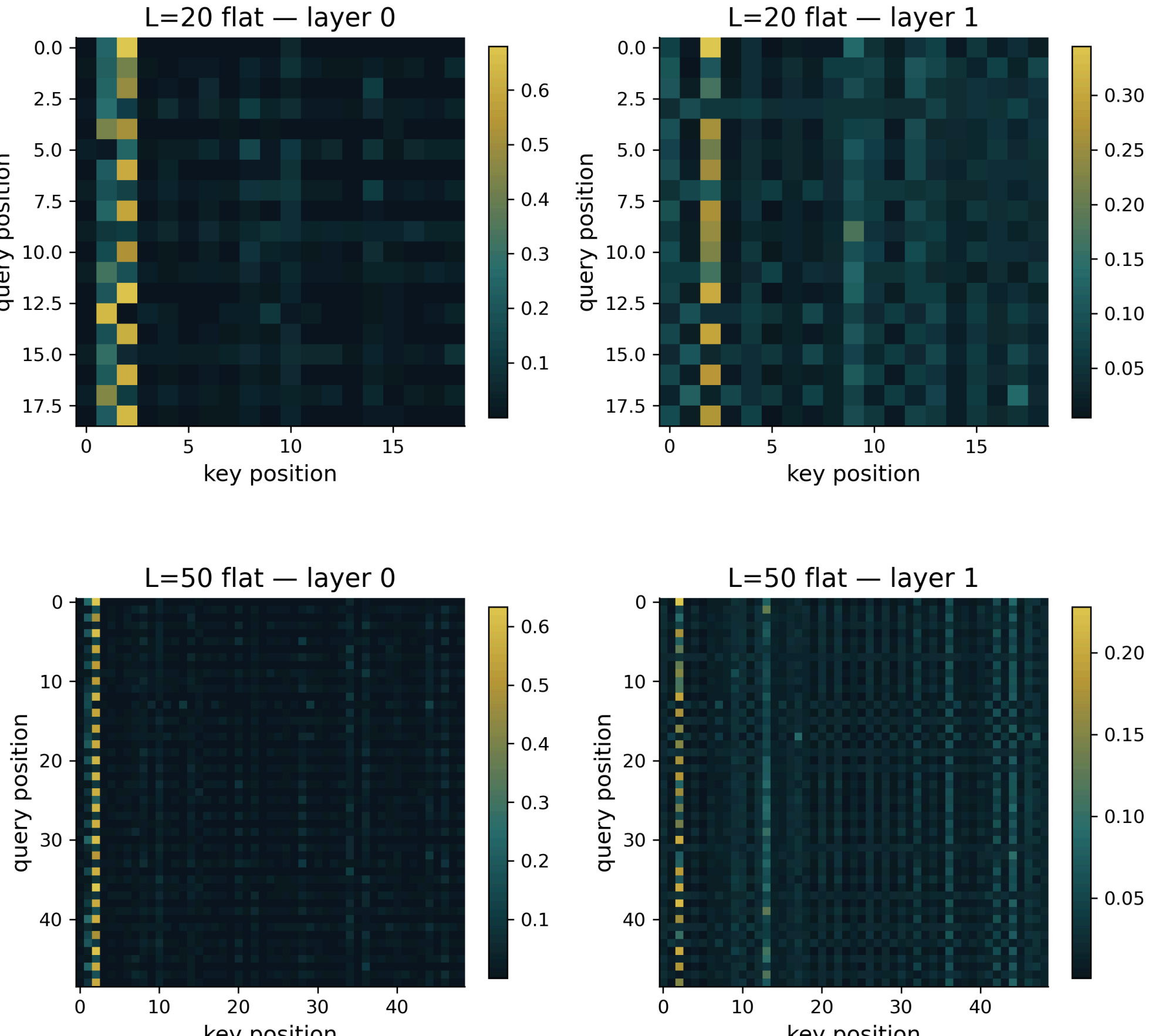

Figure 8: Attention weight matrices for TF $d = 64$ (layer 0 and layer 1) on several probe inputs. Instead of a local band around the diagonal, weights concentrate on a few fixed key positions near the start of the sequence, a pattern known as *attention sink* [34].

If what $d = 64$ has learned were a kind of local processing similar to the NCA's, with attention weights concentrated between adjacent tokens, then it might share a mechanistic basis with the NCA, and the main paper's argument about "local geometry" as an inductive bias would no longer be a necessary condition for the emergence of Proto-CKY. We test this hypothesis by inspecting the attention weight matrices.

For a range of representative expressions (the flat `id + id * id`, flat extensions of various lengths, constructions of nesting depth 2 and 4), we extract the attention matrices from both layers of $d = 64$. If weights were concentrated along the diagonal, we should see a band structure around it.

That is not what we see (Figure 8). All attention matrices display an **aggregation column** pattern: a few specific key positions (typically among the first tokens of the sequence) absorb almost all attention from every query, while the remaining positions are nearly ignored. Every token sends its information to the same few sink points and reads an answer back from them.

To quantify this observation, we define a locality score: for each attention matrix, we compute the attention-weighted mean distance between query and key, normalized by sequence length. Strictly diagonal attention yields roughly 0; uniform attention yields roughly 1/3. The scores we measure lie between 0.28 and 0.38, not only non-local but slightly more dispersed than uniform (because the weights concentrate on aggregation columns far from the diagonal).

This mechanism is known in the Transformer literature as an attention sink. $d = 64$ has not degenerated into an implicit NCA; it has learned an architecturally different aggregation-routing strategy. This answers the question of this section: what $d = 64$ does is not the same thing.

### A.3 Depth generalization: both architectures fail, both are rescued by deep augmentation

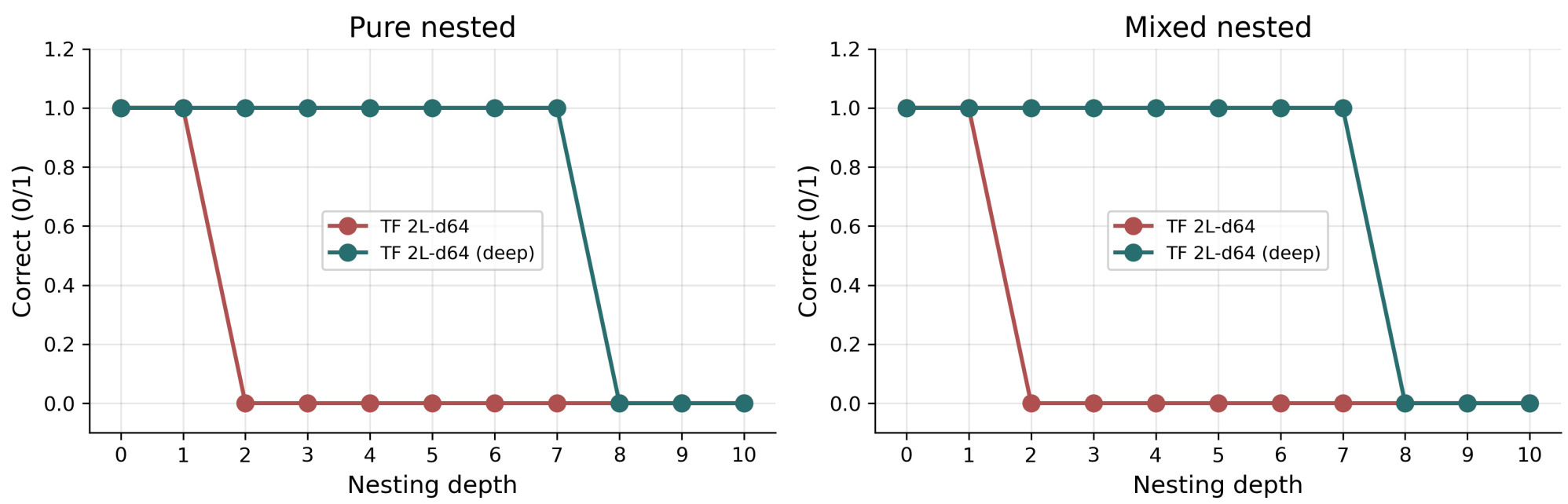

Figure 9: Depth generalization for TF $d = 64$, standard and deep-augmented. Standard training handles only depths 0 and 1; the deep-augmented variant reaches depth 7 on pure nested inputs but loses length generalization (see text). The pattern (standard fails, deep-augmented rescues depth at the cost of length) mirrors what the main NCA shows in Figure 2.

The second contrast is more decisive. We test $d = 64$ on pure nested inputs $(((\ldots(\text{id})\ldots)))$ and on mixed depth-length inputs, with depth 0 through 10. Under standard training, $d = 64$ is correct on depths 0 and 1 (within the training distribution) and fails from depth 2 onward. The same model achieves 100% accuracy on flat inputs at $L = 1000$ and 0% on the simple nesting $((\text{id}))$ at $L = 5$.

This matches the behavior of the standard NCA reported in the main paper (Figure 2) exactly. Neither architectural family achieves depth generalization under standard training.

We therefore also train a deep-augmented variant of $d = 64$, explicitly raising the nesting depth in the training distribution, following the same deep-augmentation protocol as the main NCA. The deep-augmented $d = 64$ handles pure nesting up to depth 7, comparable to the depth coverage of the deep-augmented NCA. But this variant also shows a marked regression on length: accuracy drops to around 74.5% at $L = 100$, 82% at $L = 500$, and 75% at $L = 1000$ (Figure 9).

Both architectural families, then, exhibit the same trade-off: under a single set of weights, one can do well on length or on depth, but not both. The trade-off is therefore more likely a property of the training-protocol family (on-the-fly generation, an $L \leq 12$ window, and the current loss structure) than a property of any particular architecture.

### A.4 Two kinds of success, one shared limitation

Putting the observations together:

- Transformer $d = 64$ achieves length generalization given enough parameters, but through a **non-local** aggregation mechanism (attention sink). The NCA achieves the same length generalization through a **local wavefront** mechanism. The two have no recognizable isomorphism at the mechanism level, even though they agree at the behavioral level.
- Depth generalization requires explicit augmentation of the training data, impartially across both architectural families. Proto-CKY-style hierarchical organization does not emerge "automatically"; it requires a training distribution that covers hierarchical structure.
- The length-vs-depth trade-off holds across both architectural families. This limitation is more likely tied to the current training protocol than to any particular architectural choice.

These observations do not weaken the main paper's claim; they clarify it. The main paper does not claim that "the NCA achieves something the Transformer cannot." A sufficiently large Transformer achieves the same behavioral-level generalization. The claim is something narrower and cleaner: **when this behavior is observed on a minimal 18k-parameter, purely-local-interaction system, what does its internal form look like?** The answer is an ordered grid, aligned with the geometry of the CKY chart and filled by a local wavefront. The Transformer's form of success on the same task is different: it lies not on the geometry of the CKY chart but on the topology of attention sinks. Both architectures can "get it right"; only one produces Proto-CKY.

There is a deeper distinction worth noting. The NCA's Proto-CKY is a product of **emergence**: what is trained is the local rule, and the global structure is a spontaneous consequence of iterating that rule, never directly optimized. The Transformer's attention sink is a product of **learning**: gradient descent directly shapes the weights of every layer, and the internal organization is a direct output of the optimization process. Both are "internal structures that appear after training," but their generative mechanisms differ. Emergence grows from local rules; learning is sculpted by global optimization. This distinction echoes the operational definition of emergence in the main paper (§2): emergence requires local interaction without external coordination, and the Transformer's global attention does not satisfy this constraint.